# Modelling Sentiment Analysis:
# LLMs and augmentation techniques

Guillem Senabre Prades,
*EE department, **DSML Course

**Abstract** This paper provides different approaches for a binary sentiment classification on a small training dataset. LLMs that provided state-of-the-art results in sentiment analysis and similar domains are being used, such as BERT, RoBERTa and XLNet. The reader can also find different data augmentation techniques to deal with the small amount of training data provided.

Sentiment analysis aims to identify and extract subjective information from textual data, classifying it into different sentiments, such as positive, negative, neutral or even more categories. Understanding the sentiment behind user-generated content has become crucial for businesses, organizations, and researchers, as it provides valuable insights into public opinion, customer satisfaction and brand perception.

Not long ago, sentiment analysis relied on conditional rule-based approaches, lexicons, and simple machine learning algorithms. However, the growth of deep learning and the introduction of Large Language Models (LLMs) have revolutionized the field, enabling more accurate and sophisticated sentiment analysis techniques. LLMs such as BERT, RoBERTa (Liu et al., 2019) and XLNet (Yang et al., 2020) have demonstrated remarkable performance on various NLP tasks, including sentiment analysis.

Even though LLMs or Pretrained Language Models (PTMs) have been proved to be a powerful tool to deal with NLP tasks, they need to be further trained to be domain-specific oriented with specific training datasets in that domain. Retraining the PLM is called finetuning, and there are many techniques to achieve this.

If the training dataset is small, finetuning a PLM cannot be worth it, as it requires some time and resources to do it, and the improvements may not be the expected ones. Therefore, trying to find larger datasets or mining new data could be a solution, although this is not always a possibility. That is why recurring data augmentation techniques such as "*nlpaug*" and synthetic data generation with GANs or other PLM can be a good way to go.

Motivated by the advancements in LLMs, this study aims to provide an exploring analysis between different PLM such as BERT, RoBERTa and XLNet to approach a binary sentiment classification task with a small training dataset, containing 2000 values, and a prediction set with 11000 values. Additionally, the impact of data augmentation techniques is also explored, specifically with different techniques provided by the "*nlpaug*" library and creating synthetic data with GPT-2. By evaluating the performance of each model on Kaggle, this study seeks to provide insights on the strengths and drawbacks as well as potential areas of improvement for each algorithm.

The paper will be divided in multiple sections, showing the dataset in section 1, the methodology or experimental flow in section 2, a survey on the different LLMs as well as the augmentation techniques used for the section 3, the presentation of the results in section 4, discussion, and a summary of the key concepts in section 5 and in section 6 the reader will find conclusion of the study.

## Dataset

We have been provided with a training dataset and a validation dataset. The training dataset has two features, "TEXT" with 2000 sentences and "LABEL" containing the positive ('1') or negative ('0') sentiment of the sentences. Furthermore, the length of the sentences doesn't surpass 50 tokens.

The validation set contains 11000 sentences to be predicted in one single column named "TEXT".

The negative difference on the amount of data between datasets will be approached using data augmentation techniques.

## Methodology

Since the arrival of Transformers (Vaswani et al., 2017) and PLMs such as BERT (Devlin et al., 2019), the field of sentiment analysis has noticeably grown.

This is why in this study, the pipeline will be LLMs and its goal to find the most adequate for a specific task, while exploring finetuning and data augmentation techniques.

In this paper, mostly RoBERTa (Liu et al., 2019) and XLNet (Yang et al., 2020) are used to deal with the binary classification task since they've been proved to perform better than BERT in most cases.

As a baseline, RoBERTa and XLNet will be imported using the simpletransformers library (https://simpletransformers.ai/). This library provides many models ready to be used in NLP tasks and a very friendly and easy to use framework to implement them.

As mentioned before, the training dataset contains 2000, a small amount of data compared to the 11000 sentences to be predicted. To finetune a PLM, one can think that the greater the amount of specific-domain data the greater the performance, and in a way, that is true. But, if we lack of consistent and good data, it needs to be created as accurate as possible as the original one. Therefore, in this paper two data augmentation techniques are used, very distant from one another, but serve the same purpose. Nlpaug is a library for data augmentation that provides 15 techniques to do so, although only one of these will be used in this paper, that is synonym replacement, since it's been proved to be the most effective one (Thakur et al., 2021). The other technique employs GPT-2 as a creator of synthetic data given the original data.

Since the computational and time consumption needed to re-train a LLM to fit a specific task is quite high, some techniques can be applied to reduce these issues. This experiment is going to use fine tuning and soft prompting (Qin and Eisner, 2021). Soft prompt aims to guide the PLM in the task we want it to do in the form of a prompt. This prompt will be concatenated at the beginning of each sentence and together will be used to finetune the model, giving a more precise idea of what should the output be. The experiment has been carried out both with multiple prompts and single prompts.

To be able to make comparisons, RoBERTa and XLNet have been tested out also without data augmentation and nor soft prompts. Furthermore, when using these techniques, different values such as number of augmentations and quality and quantity of soft prompts have been tested as well.

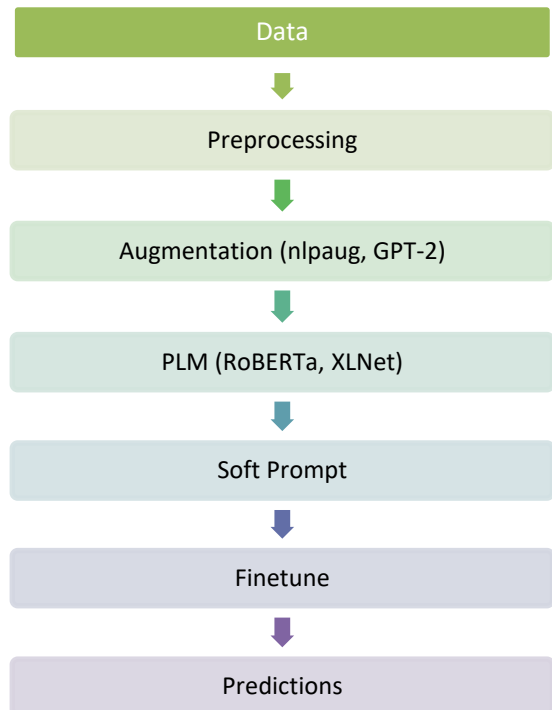

*Figure 1. Experimental flow*

Once listed the tools and methods that will be used in the study, the reader can see below the experimental flow that will be followed to obtain the results.

Before augmenting the data, a basic preprocessing is set so the data augmented is already clean. That reduces the computational power and time needed later to increase the dataset, since many tokens, like stop words, are deleted.

### Results & Discussion

Since the quantity of training data was small, all of it has been used to finetune the models. The meaning of this is that there has been no data split into training and test set. Therefore, the metric used in this experiment to evaluate the performance of the model is the accuracy given by Kaggle once the validation set is predicted. Only with this metric we can compare all the models used and see who is performing better on the validation dataset. Since many augmentations have been tested as well as many prompts and hyperparameter values, a table for each PLM test will be shown below.

|                | PLM       |         |
|----------------|-----------|---------|
| **Techniques** | RoBERTa-L | RoBERTa |
| Default        | 0.74204   | 0.70113 |
| HP testing     | 0.74204   | 0.74863 |
| nlpaug = 1     | 0.75363   | -       |
| nlpaug = 2     | 0.77659   | -       |
| nlpaug = 3     | 0.75340   | -       |
| nlpaug = 4     | 0.76931   | -       |
| SP = 1         | 0.7775    | -       |
| SP = 7         | 0.77181   | -       |
| GPT = 2 + SP   | 0.7834    | -       |
| GPT = 4 + SP   | 0.72863   | -       |
| GPT+nlpaug+SP  | 0.7809    | -       |

*Table 1. RoBERTa outcomes with different parameters and techniques*

|                | PLM     |
|----------------|---------|
| **Techniques** | XLNet   |
| Default        | -       |
| nlpaug = 3     | 0.73886 |
| nlpaug = 16    | 0.7184  |
| GPT = 2        | 0.7209  |

*Table 2. XLNet outcomes with different techniques and parameters*

XLNet provided state-of-the-art results in other studies (Pipalia et al., 2020) but did not perform as well as BERT related models in this one. Different amounts of augmentations have been tried in both models, but when augmenting over 2 times, the performance does not improve and even decreases.

After trying GPT for data augmentation without Soft Prompting, the results were not better than nlpaug with 2 augmentations, although after mixing GPT augmentations and soft prompting, the result was the best one obtained. On the other side, when mixing both nlpaug and GPT augmentations, the result was not better.

Nlpaug provides uses synonym replacement while GPT-2 tries to imitate what he has already seen, so it is supposed to be more accurate, although using 4 augmentations with GPT-2 did not improve the results as well.

Furthermore, the performance of the models increased with one single soft prompt, but increasing the number of prompts did not increase the accuracy of the model.

Finally, we want to point out that many hyperparameter's values have been tested out and the results shown above are the output of the best combination of them, mixed with the listed techniques.

## Conclusion

This study has employed different LLM and finetuning techniques as well as data augmentation methods with a small training dataset to observe the performance on a binary sentiment classification.

It has been observed that soft prompting and data augmentation techniques, both with nlpaug and GPT-2 were useful although if the data was too much increased or more than one prompt was used, the performance was negatively affected.

Finally, it needs to be point out that there is still a long way of improvement, since other LLM and techniques could have been used, but it was an excellent way of finding new ways of dealing with text and how to apply them.

## Comments

The code and datasets used, developed and commented in this paper can be found in the following GitHub repository: https://github.com/BakiRhina/Enhancing-Sentiment-analysis-with-LLM-and-data-augmentation-techniques

## References

Devlin, J., Chang, M.-W., Lee, K., Toutanova, K., 2019. BERT: Pre-training of Deep Bidirectional Transformers for Language Understanding. https://doi.org/10.48550/arXiv.1810.04805

Liu, Y., Ott, M., Goyal, N., Du, J., Joshi, M., Chen, D., Levy, O., Lewis, M., Zettlemoyer, L., Stoyanov, V., 2019. RoBERTa: A Robustly Optimized BERT Pretraining Approach.

Pipalia, K., Bhadja, R., Shukla, M., 2020. Comparative Analysis of Different Transformer Based Architectures Used in Sentiment Analysis, in: 2020 9th International Conference System Modeling and Advancement in Research Trends (SMART). Presented at the 2020 9th International Conference System Modeling and Advancement in Research Trends (SMART), pp. 411–415. https://doi.org/10.1109/SMART50582.2020.9337081

Qin, G., Eisner, J., 2021. Learning How to Ask: Querying LMs with Mixtures of Soft Prompts.

Thakur, N., Reimers, N., Daxenberger, J., Gurevych, I., 2021. Augmented SBERT: Data Augmentation Method for Improving Bi-Encoders for Pairwise Sentence Scoring Tasks.

Vaswani, A., Shazeer, N., Parmar, N., Uszkoreit, J., Jones, L., Gomez, A.N., Kaiser, L., Polosukhin, I., 2017. Attention Is All You Need.

Yang, Z., Dai, Z., Yang, Y., Carbonell, J., Salakhutdinov, R., Le, Q.V., 2020. XLNet: Generalized Autoregressive Pretraining for Language Understanding.